\documentclass[runningheads]{llncs}
\usepackage[T1]{fontenc}
\usepackage{graphicx}
\usepackage{booktabs}
\usepackage[misc]{ifsym}
\newcommand{\corr}{(\Letter)}
\usepackage[labelformat=simple]{subcaption}
\usepackage{multirow}
\usepackage{multicol}    % For spanning both columns
\usepackage{appendix}

\usepackage[utf8]{inputenc} % allow utf-8 input
\usepackage{hyperref}       % hyperlinks
\usepackage{url}            % simple URL typesetting
\usepackage{nicefrac}       % compact symbols for 1/2, etc.
\usepackage{microtype}      % microtypography
\usepackage{xcolor}         % colors

\usepackage{svg}

\usepackage{mathrsfs}
\usepackage{amsfonts}       % blackboard math symbols
\usepackage{amsmath}
\usepackage{amssymb}

\usepackage{bbold}
\usepackage{mathtools}
\usepackage{enumerate}
\usepackage{pgfplots}
\usepackage{wrapfig}

\newcommand{\R}{\mathbb{R}}

\pgfplotsset{compat=1.18} 
\begin{document}

\title{MixerFlow: MLP-Mixer meets Normalising Flows}

%\titlerunning{Underwater Basket Weaving Under Extreme Pressure}
% If the full title of your paper is short enough to also fit in the running head, you can omit the abbreviated paper title here. You can check as follows: if you comment out the \titlerunning line, something will appear in the header of all odd-numbered pages of your PDF from page 3 onward. This something is either the full title (in which case all is well), or the error message "Title Suppressed Due to Excessive Length". If this error message appears, you're going to want to provide an abbreviated title within the \titlerunning command, because if you won't do it, Springer will do it for you.

%N.B.: Author information (both in the \author{} and \authorrunning{} command) should only be present in the Camera-Ready Version of your paper. The version that you initially submit for review, ought to be double-blind. So, when initially submitting your paper, use:
%\author{Author information scrubbed for double-blind reviewing}
\author{Eshant English \inst{1} \corr \and
Matthias Kirchler\inst{1,2,3}  \and
Christoph Lippert\inst{2,3}}
% You may leave out the orcidID information, if you want to.
% Use \corr to indicate the corresponding author. Note the spacing around the \corr command. Only one author can be the corresponding author.

%N.B.: comment out the \authorrunning{} command for the double-blind version of your paper submitted for review. Later, if your paper is accepted, use the command for the Camera-Ready Version.
\authorrunning{E. English et al.}
% If there are two authors, write 'A.L. Benjamin and C.C. Broadus Jr.'
% If there are more than two authors, '[...] et al.' is used.

\institute{Hasso Plattner Institute for Digital Engineering, Germany \email{\{eshant.english\}@hpi.de}
\and
University of Kaiserslautern-Landau, Germany 
\and
Hasso Plattner Institute for Digital Health at the Icahn School of Medicine at Mount Sinai, NYC, USA
}

\tocauthor{Eshant English, Matthias Kirchler, Christoph Lippert} 
\toctitle{MixerFlow: MLP-Mixer meets Normalising Flows}
\maketitle              % typeset the header of the contribution

\begin{abstract}
Normalising flows are generative models that transform a complex density into a simpler density through the use of bijective transformations enabling both density estimation and data generation from a single model. %However, the requirement for bijectivity imposes the use of specialised architectures.
  In the context of image modelling, the predominant choice has been the Glow-based architecture, whereas alternative architectures remain largely unexplored in the research community. In this work, we propose a novel architecture called MixerFlow, based on the MLP-Mixer architecture, further unifying the generative and discriminative modelling architectures. MixerFlow offers an efficient mechanism for weight sharing for flow-based models. Our results demonstrate comparative or superior density estimation on image datasets and good scaling as the image resolution increases, making MixerFlow a simple yet powerful alternative to the Glow-based architectures. We also show that MixerFlow provides more informative embeddings than Glow-based architectures and can integrate many structured transformations such as splines or Kolmogorov-Arnold Networks.

\end{abstract}

\section{Introduction}

Normalising flows \cite{papamakarios2021normalizing,Kobyzev_2021}, a class of hybrid statistical models, serve a dual purpose by functioning as both density estimators and generative models. They achieve this versatility through a series of invertible mappings, enabling efficient inference and generation from the same model. One of their distinctive features is explicit likelihood training and evaluation, distinguishing them from models relying on lower-bound approximations \cite{kingma2022autoencoding,ho2020denoising}. Furthermore, normalising flows offer computational efficiency for both inference and sample generation, setting them apart from Autoregressive models like PixelCNNs \cite{oord2016conditional}.

The invertible nature of normalising flows extends their utility to various domains, including solving inverse problems \cite{peters2019generalized} and enabling significant memory savings during the backward pass, where activations can be efficiently computed through the inverse operations of each layer. Additionally, normalising flows can be trained simultaneously on a supervised prediction task and the unsupervised density modelling task to function as generative classifiers or in the context of semi-supervised learning \cite{nalisnick2019hybrid}, setting them apart from other generative models such as GANs \cite{goodfellow2014generative} and Diffusion models \cite{ho2020denoising}.

Despite their wide-ranging applications, normalising flows lack expressivity. The Glow-based architecture \cite{kingma2018glow} has become the standard for implementing normalising flows due to its clever design, often requiring an excessive number of parameters, especially for high-dimensional inputs. Existing literature primarily focuses on enhancing expressiveness, employing strategies such as coupling layers with spline-based transformations \cite{durkan2019neural}, kernelised layers \cite{english2023kernelized}, log-CDF layers \cite{ho2019flow++}, or introducing auxiliary layers like the Butterfly layer \cite{pmlr-v162-meng22a} or 1x1 convolution \cite{kingma2018glow}.

Nevertheless, alternative architectures remain relatively unexplored, with ResFlows \cite{chen2020residual} being a notable exception. In this work, we introduce MixerFlow, drawing inspiration from the MLP-Mixer \cite{tolstikhin2021mlpmixer}, a well-established discriminative modelling architecture. Our results demonstrate that MixerFlow consistently performs well, matching or surpassing the negative log-likelihood of the widely adopted Glow-based baselines. MixerFlow excels in scenarios involving uncorrelated neighbouring pixels or images with permutations and scales well with an increase in image resolution, outperforming the Glow-based baselines. Furthermore, our experiments suggest that MixerFlow learns more informative representations than the baselines when training hybrid flow models and can easily integrate other transformations such as Spline \cite{durkan2019neural} or Kolmogorov-Arnold Networks(KAN) layer \cite{liu2024kan} with increased expressiveness.
%Our work opens up new avenues for exploring alternative architectures for normalising flows, with MixerFlow showing promising potential.

\section{Related Works}
Our work is closely related to the MLP-Mixer architecture \cite{tolstikhin2021mlpmixer}. However, the MLP-Mixer is designed for discriminative tasks and lacks inherent invertibility, a critical requirement for modelling flow-based architectures.

The field of flow models is quite extensive, including well-known models such as Glow \cite{kingma2018glow}, Neural Spline Flows \cite{durkan2019neural} integrating splines into coupling layer, Generative Flows with Invertible Attention \cite{sukthanker2022generative} replacing convolutions with attention, Butterfly Flow \cite{pmlr-v162-meng22a}, augments a coupling layer with a butterfly matrix, and Ferumal Flows \cite{english2023kernelized}, kernelises a coupling layer. These models often share a foundational Glow-like architecture, with specific component modifications aimed at improving inter-data communication and mixing for an improved expressivity in normalising flows in different ways. MixerFlow provides an alternative architecture to Glow \cite{kingma2018glow} with the possibility to add the specific components from these works to improve inter-data communication. %For instance, \citet{sukthanker2022generative} applied attention instead of convolutions in the Glow architecture and \citet{english2023kernelized} used kernels to replace traditional neural networks.

Another noteworthy approach is the Residual-Flow-based framework \cite{chen2020residual}, which encompasses Monotone Flows \cite{ahn2022invertible} and ResFlows. In contrast to Glow-based architectures, ResFlow ensures invertibility through fixed-point iteration \cite{behrmann2019invertible}. These two categories, Glow-based and ResFlow-based, exhibit distinct characteristics and can be considered the primary classes of flow-based architectures for image modelling.

In addition to these, there exist other flow methods such as Gaussianisation Flows \cite{meng2020gaussianization} within the Iterative Gaussianisation \cite{Laparra_2011} family and FFJORD \cite{grathwohl2018ffjord} in the family of continuous-time normalising flow. These methods represent different methodological approaches to flow-based modelling, adding richness to the landscape of techniques available for density estimation. 

Our proposed architecture leverages the inherent weight-sharing properties of the MLP-Mixer. This design choice allows for flexibility in integrating either a coupling layer, a Lipschitz-constrained layer \cite{behrmann2019invertible} commonly found in ResFlow-like architectures, or even an FFJORD layer. This ensures that our model can leverage any flow method providing versatility across various application scenarios.

Our attempt to further unify generative and discriminative architectures finds resonance in similar attempts within the research community. For instance, VitGAN \cite{lee2021vitgan} adapts the VisionTransformer \cite{dosovitskiy2021image} architecture for generative modelling, demonstrating the adaptability of existing architectures. Another noteworthy example is by \cite{nalisnick2019hybrid}, which leverages generative architecture for discriminative tasks in a hybrid context.

\section{Preleminaries}

In this section, we briefly introduce the major components of a normalising flow and the MLP-Mixer architecture.
\paragraph{\textbf{The change of variables theorem}} states that if  $p_X$ is a continuous probability distribution on $\R^d$, and $f : \R^d \to \R^d$ is an invertible and continuously differentiable function with $z := f(x)$, then $p_X(x)$
%and  $p_Z$ is the chosen and known distribution such that $z = f({x})$ is an invertible transformation of $x$. Then the density at $p_X({x})$
can be computed as
\begin{align*}
    p_X({x}) = p_Z(z) \left\vert \text{det}\left(\frac{\partial f({x})}{\partial {x}}\right) \right\vert.
\end{align*}
If we model $p_Z$ with a simple parametric distribution (such as the standard normal distribution), then this equation offers an elegant approach to determining the complex density $p_X({x})$, subject to two practical constraints. First, the function $f$ must be bijective. Second, the Jacobian determinant should be readily tractable and computable.

\paragraph{\textbf{Non-linear coupling layers:}}
Coupling layers represent an essential component within the framework of most flow-based models. These non-linear layers serve a pivotal role by enabling efficient inversion for normalising flows whilst maintaining a tractable determinant Jacobian. Various forms of coupling layers exist, including additive and affine, amongst others. Below, we briefly define the affine coupling layers

\paragraph{\textbf{Affine coupling layers}} involve splitting the input $x$ into two distinct components: $x_a$ and $x_b$. Whilst the first component ($x_a$) undergoes an identity transformation, the second component ($x_b$) undergoes an affine transformation characterised by parameters $S$ and $T$. These non-linear parameters are acquired through learning from $x_a$ using a function approximator, such as neural networks. The final output of an affine coupling layer is the concatenation of these two transformations. The equations below summarise the operations.

\[X^{d}, X^{D-d} = \text{split}(X)\]
\[S, T = F(X_{D-d})\]
\[
Y_{D-d} = X_{D-d}
\]
\[
Y_{d} = S \odot X_{d} + T
\]
Given that only one partition undergoes a non-trivial transformation within a coupling layer, the choice of partitioning scheme becomes crucial. It has been shown that incorporating invertible linear layers can aid in learning an enhanced partitioning scheme \cite{kingma2018glow,pmlr-v162-meng22a}.

\paragraph{\textbf{MLP-Mixer:}} The MLP-Mixer \cite{tolstikhin2021mlpmixer} is an architecture tailored for discriminative vision tasks, relying exclusively on multi-layer perceptrons (MLPs) for its operations. It distinctively separates per-location and cross-location operations, which are fundamental in deep vision architectures and often co-learnt in models like Vision Transformers (ViT) \cite{dosovitskiy2021image} and Convolutional neural networks \cite{oshea2015introduction} with larger kernels.

The central concept of this architecture begins with the initial partitioning of an input image into non-overlapping patches, denoted as $n_p$. If each patch has a resolution of $(p_h, p_w)$, and the image itself has dimensions $(h, w)$, then the number of patches is calculated as $n_p = (h * w)/(p_h * p_w)$. Each patch undergoes the same linear transformation, projecting them into lower dimensional fixed-size vectors represented as $c$. These transformed patches collectively form a new representation of the input image in the form of a matrix, we refer to it as the ``mixer-matrix,'' with dimensions $n_p \times c$. Here, $c$ corresponds to the dimensionality of the projected patch, often referred to as ``channels'' in MLP-Mixer literature, and $n_p$ represents the total number of patches, typically called ``tokens.''

The core innovation of the MLP-Mixer unfolds through the application of multi-layer perceptrons (MLPs), which are applied twice in each mixer layer. The first MLP, known as the token-mixing MLP, operates on columns of the mixer-matrix. The same MLP is applied to all columns of the mixer-matrix. The second MLP referred to as the channel-mixing MLP, is applied to the rows of the mixer-matrix, again with the same MLP applied across all columns of the mixer-matrix. This design choice ensures weight sharing within the architecture. The mixer layer is repeatedly applied for several iterations, facilitating complete interactions between all dimensions within the image matrix, a process aptly referred to as ``mixing.''

In addition to these operations, two critical components are integral to the MLP-Mixer architecture. Firstly, layer normalisation \cite{ba2016layer} is employed to stabilise the network's training dynamics. Secondly, skip connections \cite{he2015deep} are introduced after each mixing layer to facilitate the flow of information and enable smoother gradient propagation throughout the network. An important property of the MLP-Mixer architecture is that the hidden widths of token-mixing and channel-mixing MLPs are independent of the number of input patches and the patch size, respectively, making the computational complexity of the model linear in the number of patches, in contrast to ViT \cite{dosovitskiy2021image}, where it is quadratic.

\section{MixerFlow architecture and its components}
In our architectural design, we first apply a $1\times 1$ convolution to the RGB channels of the input image with a resolution of $(h, w)$. This results in a transformed representation of the image in which the RGB channels are no longer distinguishable. Subsequently, we partition this transformed view into non-overlapping patches (or stripes, bands, dilated patches), denoted as $n_p$, each with a resolution of $(p_h, p_w)$. The choice of patch resolution is made to achieve the desired granularity, ensuring that $n_p = (h * w) / (p_h * p_w)$.
These small patches are then flattened into vectors of size $c = p_h * p_w * 3$, yielding the mixer-matrix of dimensions $n_p \times c$.

Next, we introduce two distinct types of normalising flows: channel-mixing flows and patch-mixing flows, resembling operations similar to channel-mixing MLPs and token-mixing MLPs respectively. Channel-mixing flows facilitate interactions between different channels by processing individual rows of the mixer-matrix, operating on each patch independently. Conversely, patch-mixing flows focus on interactions between different patches, processing individual columns of the mixer-matrix, whilst operating on each channel separately. These two flow operations are executed iteratively in an alternating fashion, enabling interactions between all elements within the mixer-matrix.

\begin{figure}
    
    \centering
    \includegraphics[width=1.0\linewidth]{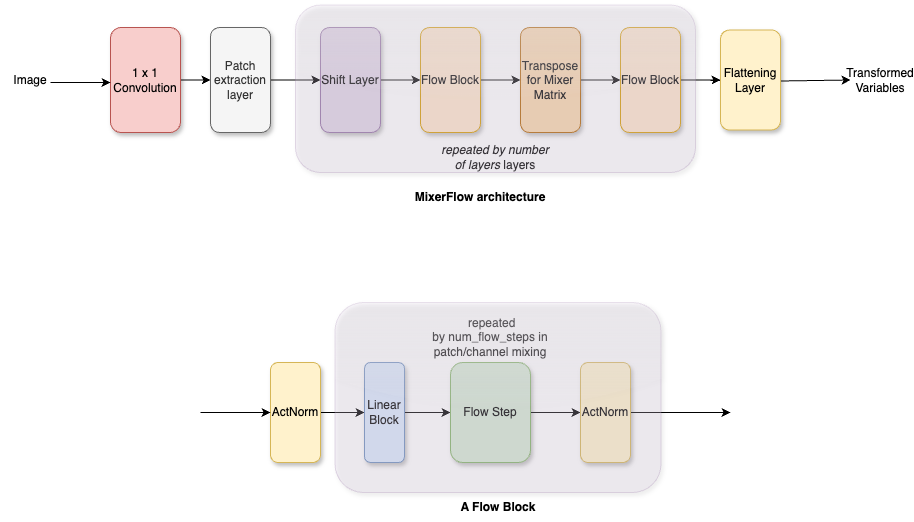}
    \caption{MixerFlow architecture}
    \label{fig:architecture}
\end{figure}

In summary, we apply the same channel flow to all rows of the mixer-matrix and the same patch flow to all columns of the mixer-matrix. This configuration ensures the desired parameter sharing across the model. These flows, both channel-flows and patch-flows, are seamlessly integrated into our architecture, with each comprising a series of subsequent components. (see Figure~\ref{fig:architecture} for an illustration)

\paragraph{\textbf{Shift layer:}}
%This layer represents a significant enhancement to the traditional mixer architecture within our MixerFlow framework.
In the MLP-Mixer, the patch definition remains static across the entire architecture, albeit projected into a lower dimension $c$ at the outset—an operation constrained by invertibility concerns.
Whilst this approach suits the MLP-Mixer, it introduces grid-like artefacts when applied to sampled images in our MixerFlow model.

To address this issue, shift layers reshape the mixer-matrix back into its original image dimensions.
It then creates a frame, leaving the inner part as $(s_h:h-p_h+s_h,s_w:w-p_w+s_w)$, where $(p_h , p_w)$ denotes the patch resolution, $(s_h , s_w)$ denotes the shifting unit, and $(h , w)$ corresponds to the image resolution.
Subsequently, we re-transform the inner part into an input resembling a mixer-matrix, which we refer to as the ``shifted-mixer-matrix.'' This process effectively shifts the patch extraction by $(s_h , s_w)$ units both vertically and horizontally respectively compared to the old patch definition.
This frame carving reduces the number of patches as some input variables are omitted by extraction of the frame.
A full mixer layer is applied to the shifted-mixer-matrix, and after this layer, the carved-out frame is reintroduced for the next stage. This strategy ensures robust interactions near the boundaries.

Importantly, the alteration in patch definition has no adverse impact on performance, as neighbouring pixels exhibit a high correlation. Moreover, it facilitates the distribution of transformations for the variables within the carved-out frame, with a focus on the non-carved-out-frame layers.
In preliminary experiments we found these shift layers to result in improvements both qualitatively and quantitatively.

\paragraph{\textbf{Linear block:}} As previously mentioned, a coupling layer operates exclusively on approximately half of the input dimensions, underscoring the importance of partition selection. RealNVP \cite{dinh2017density} introduced alternating patterns, which, in certain cases, introduce an order bias.
In contrast, Glow \cite{kingma2018glow} advocated for $1\times 1$ convolutions in the form of PLU factorisation, with fixed permutation matrix $P$ and optimisable lower and upper triangular matrices $L$ and $U$, providing a more generalised approach to permutations.
In our approach, we employ either LU factorisation or RLU, where R is a permutation matrix that reverses the order of dimensions. % denoting a specialised permutation matrix that reverses the order of dimensions.
It is crucial to emphasise the necessity of a linear block that effectively reverses the shuffling before the application of a shift layer which aims to capture the interaction between the patch boundaries. Subsequently, the linear block is followed by a Flow coupling layer.

\paragraph{\textbf{Flow layer:}} After each Linear Block, we incorporate a flow layer into our architecture. Specifically, we employ a standard affine coupling layer as our chosen flow layer, applicable to either a channel flow or a patch flow. Within each flow layer, we use a residual block as the function approximator. In our experiments, we mostly chose a latent dimension of 128 for both the channel-flow-residual-block and the patch-flow-residual-block, along with the GELU \cite{hendrycks2023gaussian} activation function and batch normalisation \cite{ioffe2015batch} in the layers of the residual block.

\paragraph{\textbf{ActNorm layer:}} Due to the computation of a full-form Jacobian when applying layer normalisation \cite{ba2016layer}, we opt for ActNorm \cite{kingma2018glow} as the preferred normalisation technique in our architecture. ActNorm layers are data-dependent initialised layers with an affine transformation that initialises activations to have a mean of zero and a variance of one based on the first batch. In contrast to Glow-based architectures where ActNorm layers are applied only to the channels, we apply ActNorm to all activations after each flow layer and just before the initial linear layer following each transpose operation (i.e., applying a flow layer to columns of the matrix after applying it to the rows or vice versa).

\paragraph{\textbf{Identity initialisation:}}  We initialised all linear blocks to perform an identity function. Additionally, we initialise the final layer of each residual block within the flow with zeros to achieve an identity transformation.  This approach, as reported by \cite{kingma2018glow}, has been observed to be beneficial for training deeper flow networks.

\section{Experiments}

We conducted an extensive series of experiments encompassing various datasets, varying dataset sizes, and diverse applications. These applications include permutations, classification tasks, and the integration of Masked Autoregressive Flows \cite{papamakarios2018masked} into our MixerFlow model. We use thirty MixerFlow layers with a shift of either one or two every fourth layer in our experiments. We perform uniform dequantisation, use a patch size of four for $32\times 32$ resolution and a patch size of eight for $64\times 64$ resolution and use Adam for optimisation with the default parameters. For baseline experiments, we used the experimental settings as in \cite{lu2021woodbury}, reproducing results on standard datasets. Following standard practices in flow literature, we reported density estimation results in "bits-per-dimension" (BPD) and reported results up to two decimal places, which led to the exclusion of standard error intervals as they were consistently less than 0.004. 

\begin{figure*}
  \centering
  \subcaptionbox{\label{sub:cifar10} For CIFAR-10}{\includegraphics[width=0.3\linewidth]{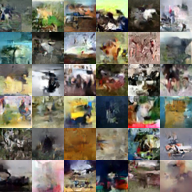}}
  \hfill
  \subcaptionbox{\label{sub:anime}For AnimeFaces}{\includegraphics[width=0.3\linewidth]{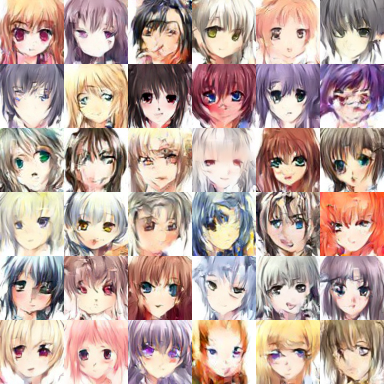}}
  \hfill
  \subcaptionbox{\label{sub:galaxy}For Galaxy32}{\includegraphics[width=0.3\linewidth]{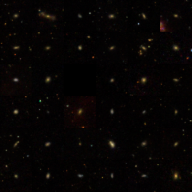}}
  \caption{Sampled images from our MixerFlow}
\end{figure*}

\subsection{Density estimation on $32\times 32$ datasets}

\paragraph{\textbf{Datasets:}} In line with previous research, we assessed the performance of MixerFlow on standard $32\times 32$ datasets, specifically ImageNet32 \cite{russakovsky2015imagenet,van2016pixel}\footnote{Two versions of downscaled ImageNet exist. The one used to evaluate normalising flow models has been removed from the official website but remains accessible through alternative sources, such as Academic Torrents.} and CIFAR-10. \cite{Krizhevsky2009LearningML}.
%It is important to note that we evaluated the older version of ImageNet, commonly used for evaluating flow-based models. This version can be downloaded from the web, for instance, via AcademicTorrents.

\paragraph{\textbf{Baselines:}} We use various Glow-based baselines for our evaluations: Glow \cite{kingma2018glow}, Neural Spline flows \cite{durkan2019neural}, MaCow \cite{ma2019macow},ME-Woodbury \cite{lu2021woodbury}, and Emerging and Periodic convolutions \cite{hoogeboom2019emerging}.

\paragraph{\textbf{Results:}}
%For a fair comparison, we keep the number of parameters for all models at roughly the same size and train all models with equivalent hyperparameter evaluation. 
Table~\ref{tab:image32} presents our quantitative results, demonstrating the competitive or superior performance of MixerFlow to all of the aforementioned baselines.
%Our scores are worse than those reported in prior works because we use smaller models and train each model on a single GPU. We also found that adding more layers can help improve the log-likelihood of our architecture.
%It is important to note that all these models, including our own, have been kept at almost the same size for the sake of fair comparisons.
The sample images from our MixerFlow trained on CIFAR-10 can be seen in Figure~\ref{sub:cifar10}.

\begin{table}[!htbp]
  \caption{Negative log-likelihood (in bits per dimension) for $32\times 32$ datasets. Smaller values are better.}
  \label{tab:image32}
  \centering
  \begin{tabular}{llll}
    \toprule
    %\multicolumn{4}{c}{Datasets}                
    %\cmidrule(r){3-7}
    
    Method & CIFAR-10     & ImageNet32 & Params \\
    \midrule
    \textbf{MixerFlow}  & \textbf{3.46} &\textbf{4.20} & 11.34M \\
    Glow &3.51  & 4.32 & 11.02M    \\
   Neural Spline &3.50 & 4.24 & \textbf{10.91M} \\
   \midrule
   MaCow     & 3.48  & 4.34 & 11.43M  \\
   Woodbury  & 3.48 & 4.22  & 11.02M \\
   Emerging  & 3.48 & 4.26  & 11.43M \\
    Periodic  & 3.49 & 4.28  & 11.21M \\
  
    %\cmidrule(r){1-7}
    \bottomrule
  \end{tabular}
\end{table}
\subsection{Density estimation on $64\times 64$ datasets}

\paragraph{\textbf{Datasets:}} We assessed the performance of MixerFlow on two distinct datasets: ImageNet64 \cite{russakovsky2015imagenet}, a standard vision dataset, and AnimeFace \cite{naftali2023aniwho}, a collection of Anime faces.

\paragraph{\textbf{Baselines:}} For our evaluations, we use a couple of Glow-based baselines, specifically Glow \cite{kingma2018glow}, and Neural Spline \cite{durkan2019neural}.

\paragraph{\textbf{Results:}} Our quantitative results in Table~\ref{tab:image64} illustrate that MixerFlow outperforms the selected baselines in terms of negative log-likelihood measured in bits per dimension. Notably, our analysis of model sizes reveals that MixerFlow scales remarkably well as image size increases, requiring approximately half the number of parameters compared to the other baselines. This outcome aligns with our expectations, as the hidden patch-flow-MLP dimension remains independent of the number of patches, and the hidden-channel-flow-MLP dimension remains independent of the number of channels—a characteristic inherited from the MLP-Mixer architecture. For a visual representation of our results, please refer to Figure~\ref{sub:anime}, which displays sample images generated by our MixerFlow model trained on the AnimeFace dataset.

\begin{table}[!htbp]
  \caption{Negative log-likelihood (in bits per dimension) for $64\times 64$ datasets. Smaller values are better.}
  \label{tab:image64}
  \centering
  \begin{tabular}{llll}
    \toprule
    %\multicolumn{4}{c}{Datasets}                
    %\cmidrule(r){3-7}
    
    Method & AnimeFaces     & ImageNet64 & Params \\
    \midrule
    Glow &3.21  & 3.94 & 37.04M    \\
    Neural Spline &3.23  & 3.95 & 38.31M    \\
  \textbf{MixerFlow}  & \textbf{3.17} & \textbf{3.92} & \textbf{18.90M}\\
    %\cmidrule(r){1-7}
    \bottomrule
  \end{tabular}
\end{table}

In the context of the AnimeFaces dataset, we also observed a qualitative improvement. Specifically, we noted a reduction in artefacts compared to the Glow-based baseline (Figure~\ref{fig: anime glow full} and Figure~\ref{fig: anime full} in the Appendix).

\subsection{Enhancing MAF with the MixerFlow}

Masked Autoregressive Flows (MAF) \cite{papamakarios2018masked} represent one of the most popular density estimators for tabular data. They are a generalisation of coupling layer flows, such as Glow and RealNVP. Notably, MAF tends to outperform coupling layer flows on tabular datasets, although it comes at the cost of relatively slow generation. The concept of MAF emerged as an approach to enhance the flexibility of the autoregressive model, MADE \cite{germain2015made}, by stacking their modules together. This innovation, which enables density evaluations without the typical sequential loop inherent to autoregressive models, significantly accelerated the training process and enabled parallelisation on GPUs.

\begin{figure}%{r}{0.4\linewidth}
    \centering

    \includegraphics[width=0.4\linewidth]{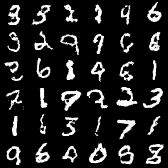}
    \caption{Sampled Images for our MAF-based MixerFlow on MNIST}
    \label{MAF-MNIST}
\end{figure}

However, MAF is vulnerable to the curse of dimensionality, necessitating an enormous number of parameters to achieve scalability for image modelling. This requirement for many additional layers can render MAF impractical for certain tasks, i.e. image modelling. In our work, we demonstrate that by substituting a flow step with an MAF layer within our MixerFlow model, we can enable MAF for density estimation tasks involving image datasets. This integration leverages the effective weight-sharing architecture inherent in MixerFlow and substantially enhances the performance of the MAF model.

\paragraph{\textbf{Datasets:}} We assessed the performance of the MAF integration on two distinct datasets: MNIST \cite{deng2012mnist} and CIFAR-10 \cite{Krizhevsky2009LearningML}.

\paragraph{\textbf{Baselines:}} We considered MAF \cite{papamakarios2018masked} and MADE \cite{germain2015made} as the baselines for our evaluation as they are pre-cursors of MAF's integration into our architecture.

\paragraph{\textbf{Results:}} Our findings, as presented in Table~\ref{tab:MAF}, showcase the density estimation results with MixerFlow further enhancing the use of the MADE module through MAF's integration in our architecture for MNIST \cite{deng2012mnist} and CIFAR-10 \cite{Krizhevsky2009LearningML}. Furthermore, Figure~\ref{MAF-MNIST} provides a visual representation of the generated samples from our enhanced MAF model on the MNIST dataset

\begin{table}[!htbp]
  \caption{Negative log-likelihood (in bits per dimension) for MAF's integration into MixerFlow. Smaller values are better.}
  \label{tab:MAF}
  \centering
  \begin{tabular}{llll}
    \toprule
    %\multicolumn{4}{c}{Datasets}                
    %\cmidrule(r){3-7}
    
    Method & MNIST     & CIFAR-10 \\
    \midrule
    MADE  & 2.04 & 5.67 \\
    MAF & 1.89  & 4.31  \\
  \textbf{Ours}  & \textbf{1.22} & \textbf{3.44} \\
    %\cmidrule(r){1-7}
    \bottomrule
  \end{tabular}
\end{table}

\subsection{Datasets with specific permutations}
Whilst Glow-based models exhibit expressive capabilities in capturing image dynamics, they heavily rely on convolution operations involving neighbouring pixels to transform the distribution. In contrast, our MixerFlows use multi-layer perceptrons as function approximators, making them invariant to changes in pixel locations within patches and patch locations in the image. This might be helpful if there is some data corruption that induces permutation as Glow-based architectures will result in poor density estimation in such cases. In this section, we empirically demonstrate this advantage across various datasets. 

\begin{figure}[!htbp]
  \centering
  \subcaptionbox{\label{sub:shuffle_glow} For Glow}{\includegraphics[width=0.45\linewidth]{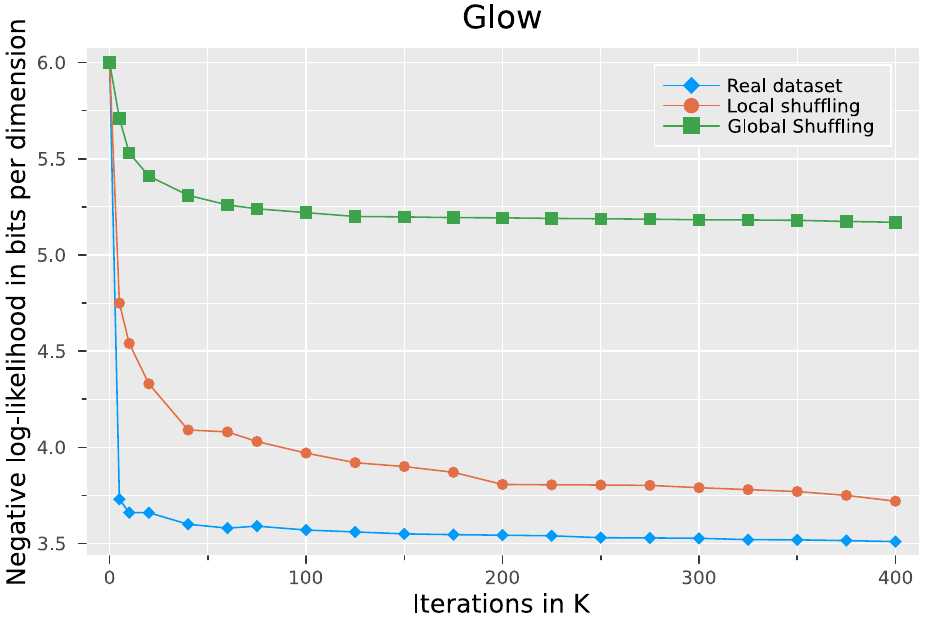}}
  \hfill
  \subcaptionbox{\label{sub:shuffle_mixer}For MixerFlow}{\includegraphics[width=0.45\linewidth]{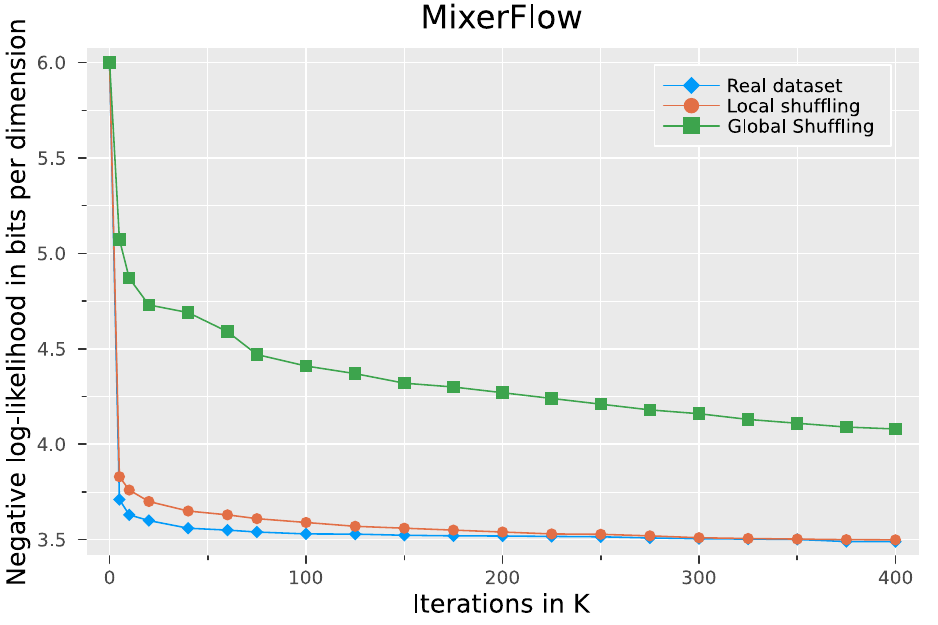}}

\end{figure}

\subsubsection{Density estimation on permuted image datasets}

\paragraph{\textbf{Dataset:}}  Our experimentation involved two types of artificial permutations. Firstly, we divided the input image into patches and performed permutations on both the order of these patches and the pixels within each patch, using a shared permutation matrix. Subsequently, we reorganised these shuffled patches in the image form, referring to this process as ``local shuffling.''
This setup is analogous to the pixel shuffling experiment introduced by \cite{tolstikhin2021mlpmixer}, but adapted to the case of density estimation. 

\begin{table}[!htbp]
  \caption{Negative log-likelihood (in bits per dimension) for the global shuffling experiment. Smaller values are better.}
  \label{tab:shuffling}
  \centering
  \begin{tabular}{llll}
    \toprule
    %\multicolumn{4}{c}{Datasets}                
    %\cmidrule(r){3-7}
    
    Global shuffling & CIFAR-10 & Imagenet32  & Params \\
    \midrule
    Glow & 5.18  & 5.27 & 44.24M    \\
    Neural Spline & 5.01  & 5.32 & 89.9M    \\
     \textbf{MixerFlow}  & \textbf{4.09} & \textbf{4.91} & \textbf{11.43M} \\
    Butterfly & 5.11 & 6.18 & - \\
 
    %\cmidrule(r){1-7}
    \bottomrule
  \end{tabular}
\end{table}

\begin{table}[!htbp]
  \caption{Negative log-likelihood (in bits per dimension) for the local shuffling experiment. Smaller values are better.}
  \label{tab:localshuffling}
  \centering
  \begin{tabular}{llll}
    \toprule
    %\multicolumn{4}{c}{Datasets}                
    %\cmidrule(r){3-7}
    
    Local shuffling & CIFAR-10 & Imagenet32  & Params \\
    \midrule
    Glow &  4.06  &  4.49 & 44.24M    \\
    Neural Spline & 3.99  & 4.54 & 89.9M    \\
  \textbf{MixerFlow}  & \textbf{3.49} & \textbf{4.24} & \textbf{11.43M} \\
    %\cmidrule(r){1-7}
    \bottomrule
  \end{tabular}
\end{table}

Secondly, we applied a ``global shuffling'', where we shuffled all the pixels of the entire image using the same permutations across all images. We created these modified versions of the CIFAR-10 \cite{Krizhevsky2009LearningML} and ImageNet32 \cite{russakovsky2015imagenet,van2016pixel} datasets, both of which are $32\times 32$ in size.

\paragraph{\textbf{Baselines:}} To assess the performance of our model, we compared it with Glow \cite{kingma2018glow} and Neural Spline flow \cite{durkan2019neural}. Additionally, we included a comparison with Butterfly Flows (results taken from existing literature) \cite{pmlr-v162-meng22a}, which are theoretically guaranteed to be able to represent any permutation matrix. All the baselines use the same Glow backbone. Notably, we evaluated larger models for the baselines in this experiment, highlighting that even with larger models, their performance lags in the presence of permutations.

\paragraph{\textbf{Results:}} We tested our hypothesis that MixerFlow is more effective when dealing with permutations in the dataset. The performance gap is substantial for both types of permutations applied to the dataset, as evident from Table~\ref{tab:localshuffling} and Table~\ref{tab:shuffling}, where we report lower negative log-likelihoods in bits per dimension (BPD). Figure~\ref{sub:shuffle_glow} and Figure~\ref{sub:shuffle_mixer} illustrate the training curve for both types of permutations on CIFAR-10 \cite{Krizhevsky2009LearningML} . Notably, stronger inductive biases in Glow-based architectures are highly dependent on the order of pixels, resulting in a significant performance drop compared to the original order of image pixels. Conversely, the performance drop is relatively minimal for MixerFlow. Whilst it was intuitively expected that there would be no performance difference on local shuffling datasets, there is a slight difference in the negative log-likelihood, this is due to our shift layer, which requires that neighbouring patches maintain the same structural positions as in the original image.

\subsubsection{Density estimation on a structured dataset: Galaxy32}

% \begin{figure}
%     \centering
%     \includegraphics[width=0.4\linewidth]{samples_galaxy.png}
%     \caption{Sampled images from our MixerFlow}
%     \label{sub:galaxy}
% \end{figure}
\paragraph{\textbf{Dataset:}} Real-world datasets often exhibit unique structures, such as permutations or periodicity. To enhance the robustness of our experiments, we applied MixerFlow to a real-world dataset—the Galaxy dataset. This dataset, curated by \cite{Ackermann_2018}, comprises images of merging and non-merging galaxies. Given that stars can appear anywhere in these images, they exhibit permutation invariance. Additionally, the dataset demonstrates periodicity, as it represents snapshots of a spatial continuum rather than individual isolated images. We downsampled the dataset to a resolution of $32\times 32$ for our experiments.

\paragraph{\textbf{Baselines:}} In our evaluation, we compared the performance of MixerFlow against two baseline models: Glow \cite{kingma2018glow} and Neural Spline \cite{durkan2019neural}.

\paragraph{\textbf{Results:}} The results, presented in Table~\ref{tab:galaxy}, clearly indicate that MixerFlow outperformed the chosen baselines. Figure~\ref{sub:galaxy} showcases samples generated by the MixerFlow model.

% \begin{table}[!htbp]
%     \begin{minipage}[t]{.45\linewidth}  
%   \centering
%   \captionof{table}{Negative log-likelihood (in bits per dimension) for the Galaxy32 dataset. Smaller values are better.}
%   \begin{tabular}{ll}
%     \toprule
%     %\multicolumn{4}{c}{Datasets}                
%     %\cmidrule(r){3-7}
%     Method & Galaxy32    \\
%     \midrule
%     Glow & 2.27   \\
%     Neural Spline&  2.25 \\
%   \textbf{MixerFlow}  & \textbf{2.22}\\
%     %\cmidrule(r){1-7}
%     \bottomrule
%   \end{tabular} 
%       \label{tab:galaxy}
%     \end{minipage}%
%     \hfill
% \begin{minipage}[t]{0.45\linewidth}
%   \captionof{table}{Classification loss and accuracy on CIFAR-10 when employing flow-based embeddings}
%   \centering
%   \begin{tabular}{lll}
%     \toprule
%     %\multicolumn{4}{c}{Datasets}                
%     %\cmidrule(r){3-7}
    
%     Method & Loss & Accuracy    \\
%     \midrule
%     Glow & 1.69  & 41.23\% \\ 
%     Neural Spline& 1.67 & 41.27\%\\
%   \textbf{MixerFlow}  & \textbf{1.54} & \textbf{45.11\%}\\
%     %\cmidrule(r){1-7}
%     \bottomrule
%   \end{tabular}
%   \label{tab:losses}
%     \end{minipage} 
% \end{table}

\begin{table}[!htbp]
  \centering
  \caption{Negative log-likelihood (in bits per dimension) for the Galaxy32 dataset. Smaller values are better.}
  \begin{tabular}{ll}
    \toprule
    %\multicolumn{4}{c}{Datasets}                
    %\cmidrule(r){3-7}
    Method & Galaxy32    \\
    \midrule
    Glow & 2.27   \\
    Neural Spline&  2.25 \\
  \textbf{MixerFlow}  & \textbf{2.22}\\
    %\cmidrule(r){1-7}
    \bottomrule
  \end{tabular} 
      \label{tab:galaxy}
    \end{table}%

\subsection{Hybrid modelling}
\cite{nalisnick2019hybrid} introduce a hybrid model that combines a normalising flow model with a linear classification head. This hybrid approach offers a compelling advantage for predictive tasks, as it allows for the computation of both $p(\text{data})$ and $p(\text{label}|\text{data})$ using a single network. This capability enables semi-supervised learning and out-of-distribution detection. However, achieving optimal results often requires joint objective optimisation or separate training for different components, as the objective function not only maximises log-likelihood but also minimises predictive error. Significantly over-weighting the predictive error term is sometimes necessary for achieving superior discriminative performance.

Since our MixerFlow is based on a discriminative architecture, namely MLP-Mixer, we posit that our model can perform better under similar predictive loss weighting.

\paragraph{\textbf{Dataset:}} We used the CIFAR-10 \cite{Krizhevsky2009LearningML} dataset for the downstream task of classification.
\paragraph{\textbf{Baselines:}} We evaluated the performance of MixerFlow embeddings against two baseline models' embeddings: Glow \cite{kingma2018glow} and Neural Spline \cite{durkan2019neural}.
\paragraph{\textbf{Results:}} In our experiments, we used our pre-trained models and added a classifier head to them. During training, we kept the flow parameters fixed whilst training the additional linear layer parameters. This setup effectively leveraged the representations learned by the flow models for downstream tasks. Our results in Table~\ref{tab:losses} indicate that the MixerFlow model exhibited lower losses compared to the chosen baselines. This suggests that MixerFlow embeddings capture more informative representations. %Consequently, our architecture demonstrates superior performance compared to Glow-backbone architectures.

\begin{table}[!htbp]
  \caption{Classification loss and accuracy on CIFAR-10 when employing flow-based embeddings}
  \centering
  \begin{tabular}{lll}
    \toprule
    %\multicolumn{4}{c}{Datasets}                
    %\cmidrule(r){3-7}
    
    Method & Loss & Accuracy    \\
    \midrule
    Glow & 1.69  & 41.23\% \\ 
    Neural Spline& 1.67 & 41.27\%\\
  \textbf{MixerFlow}  & \textbf{1.54} & \textbf{45.11\%}\\
    %\cmidrule(r){1-7}
    \bottomrule
  \end{tabular}
  \label{tab:losses}

\end{table}

\subsection{Integration of powerful architecture}
The flexible design of our MixerFlow architecture enables the seamless integration of diverse flow layers and powerful transformations. To assess the impact of these transformations on model performance, we conducted experiments while maintaining a consistent architecture depth of 30 layers and training all models for 50,000 steps.

\paragraph{\textbf{Datasets:}} We evaluated the performance of different transformations on the MNIST dataset \cite{deng2012mnist}.

\paragraph{\textbf{Models:}} We replaced the standard coupling-based MLPs in our MixerFlow model with alternative transformations, including Spline-based transformations \cite{durkan2019neural}, MAF-based transformation \cite{papamakarios2018masked}, KAN-layer \cite{liu2024kan} (using KANs instead of MLPs in standard coupling layers)

\paragraph{\textbf{Results:}} As shown in Table~\ref{tab:special transformations}, incorporating these advanced transformations within MixerFlow consistently yields improved density estimation performance compared to the baseline MLP model. This highlights the potential of leveraging specialised transformations further to enhance the expressive power of our proposed architecture. It is important to note that we used default parameter settings directly from their respective source papers due to computational constraints. We believe that fine-tuning these parameters could lead to even more significant improvements in performance.

%Our findings, as presented in Table~\ref{tab:special transformations}, hint that the density estimation with MixerFlow can be boosted further with powerful structured transformations. We would like to highlight that we have used the transformation parameters from the papers proposing the transformations and better results could be obtained with tuning the parameters, which we could not do due to computational bottleneck.

\begin{table}[!htbp]
  \caption{Negative log-likelihood (in bits per dimension) for different transformations in MixerFlow. Smaller values are better.}
  \label{tab:special transformations}
  \centering
  \begin{tabular}{llll}
    \toprule
    %\multicolumn{4}{c}{Datasets}                
    %\cmidrule(r){3-7}
    
    Method & BPD     & Params \\
    \midrule
    MLP-coupling  & 1.57 & 3.10M \\
    KAN-coupling & 1.48  & 7.45M \\
    Splines    & 1.21  & 5.28M \\
    MAF-based & 1.56  & 3.83M \\

    %\cmidrule(r){1-7}
    \bottomrule
  \end{tabular}
\end{table}

\section{Conclusion and limitations}
In this work, we have introduced MixerFlow, a novel flow architecture that draws inspiration from the MLP-Mixer architecture \cite{tolstikhin2021mlpmixer} designed for discriminative vision tasks. Our experimental results have demonstrated that MixerFlow consistently outperforms or competes comparatively (in terms of negative log-likelihood) with existing models on standard datasets. Importantly, it exhibits good scalability, making it suitable for handling larger image sizes. Additionally, the integration of MAF layers \cite{papamakarios2018masked} into our architecture showed considerable improvements in MAF, showcasing its adaptability and versatility for integrating normalising flow architectures beyond coupling layers.

We also explored the application of MixerFlow to datasets featuring artificial permutations and structured permutations, underlining its practicality and wide-ranging utility. To conclude, our work has highlighted the potential of the acquired representations in the downstream classification task on CIFAR-10, as evidenced by lower loss values. MixerFlow can also help existing MLP-Mixer architectures by providing a probabilistic approach to it.

One limitation of our proposed architecture is the absence of strong inductive biases typically associated with convolution-based flows.
This may be especially relevant if little training data is available.
%This becomes crucial when the primary goal is to generate images with pretty visualisation rather than density estimation. However, we believe that this can be accomplished by incorporating a linear convolution layer \cite{hoogeboom2019emerging, finzi2019invertible}. 
\section{Future work and broader impact}
Whilst our experimental analysis focuses on MixerFlow with MLP-based neural networks, it is crucial to note that MixerFlow's architecture is not restricted to specific neural network types or flow networks. It enables parameter sharing and can be adapted to various network architectures. This adaptability extends to the incorporation of residual flows \cite{chen2020residual}, attention layers \cite{sukthanker2022generative}, and convolutional layers, especially for larger image sizes and patch sizes, offering the potential for enhanced inductive biases and parameter sharing. Additionally, the inclusion of glow-like layers before applying the MixerFlow layers could further strengthen inductive biases.

Another promising avenue for improvement is enabling multiscale design \cite{kingma2018glow}, a well-established technique for boosting the performance of flow-based models. We leave the integration of multi-scale architecture, and stronger inductive biases for future work, believing it could further enhance the MixerFlow architecture's capabilities. We are optimistic about the potential of MixerFlow to advance flow-based image modelling, and we hope for more developments in architectural backbone research for flow models in the future.

\begin{credits}
\subsubsection{\ackname} We extend our gratitude to Arkadiusz Kwasigroch,
Sumit Shekhar, Thomas Gaertner, Noel Danz, Juliana Schneider, and Wei-Cheng Lai for their feedback on an earlier draft.  This research was funded by the HPI Research School on Data Science and Engineering, and by the European Commission in the Horizon 2020 project INTERVENE (Grant agreement ID: 101016775).

\subsubsection{\discintname}
The authors have no competing interests to declare that are relevant to the content of this article.

\end{credits}
%\bibliographystyle{splncs04}
%\bibliography{mybibliography}
%% Note that this preceding line implies that you store your BibTeX references in a file called 'mybibliography.bib'. If you instead store your references in a file with a different name, for instance 'references.bib', the preceding line should read '\bibliography{references}'. Whatever you do, DO NOT put the file name extension .bib inside the \bibliography command; this will trip up LaTeX compilers. 
%

\clearpage
\appendix
\section{Additional experimental details for our method}
%\label{sec:training details}
We employed the Adam optimiser exclusively for all our experiments. The learning rate is chosen $\in [0.001,0.0005]$, $\beta_1 = 0.90, \beta_2=0.999 $. We used Cosine decay for all experiments with the minimum learning rate equalling zero. In initial experiments, we found that 30 mixer layers with a shift layer every four layers and four flow layers for both, channel-mixing and patch-mixing, performed satisfactorily and we persisted with it for all of the experiments. We used an NVIDIA a100 GPU for training in our experiments. For more comprehensive information, please refer to Table~\ref{train details}.

\begin{table}[h!]
  \caption{Model Architectures and hyperparameters for our method. }
  \label{train details}
  \centering
  \begin{tabular}{llllllll}
    \toprule
    \multicolumn{7}{c}{Datasets}                   \\
    \cmidrule(r){3-8}
    &      & CIFAR-10     & Imagenet32 & Galaxy32 & Imagenet64 & AnimeFaces & MNIST\\
    \midrule
    & Resolution    & 32     & 32 & 32 & 64 & 64 & 28\\
    & Training Points      & 48,000     & 1,229,904 & 5,000 & 1,229,904 & 61,023 & 57,600\\
    \midrule
    &  type & coupling & coupling & coupling & coupling & coupling & AR \\
    & layers    & 30 & 30 & 30 & 30  &  30 & 30\\
     & patch size & 4  & 4 & 4 & 8 & 8 & 4 \\
    & hidden dimension     & 128 &128 & 64  & 128  &  128  & 64 \\
    & shift layer & 1 & 1 & 1  & 2 & 2 & None \\
    & Linear block & LU &LU & LU & LU &  LU  & RLU \\
   &  grad clip &5.0  & 5.0 & 5.0 & 5.0 & 5.0 & 5.0 \\
   & learning rate(lr) & 0.001 & 0.001  & 0.001 & 0.0005  & 0.0005  & 0.0005 \\
    &  $\beta_1$   & 0.90  & 0.90 & 0.90 & 0.90 &  0.90  & 0.90\\
  &   $\beta_2$     & 0.999  & 0.999& 0.999 & 0.999 &  0.99 &0.999  \\
   & training steps     & 400K  & 500K& 50K & 400K &   400K &400K \\
   & batch size     & 512 &512 & 512  & 200  &  200 & 512  \\
    \bottomrule
  \end{tabular}
\end{table}

%\vfill

\section{Bits-per-dimension}
 Bits-per-dimension(BPD) serves as a key measure for evaluating the efficiency of flow-based models. Bits-per-dimension quantifies the average number of bits needed to encode each dimension of the data distribution, thereby offering a normalised assessment of model performance. Lower BPD values indicate more effective compression and, consequently, superior modelling capabilities. It is calculated as the negative log-likelihood (in nats) divided by the product of the dimensions and log(2) [i.e., BPD = -(log $p_x$) / (h * w * c) / log(2)]. A minute difference in BPD can hold huge significance, e.g. a 0.01 BPD difference is equivalent to approximately 21.29 in nats/log-likelihood for a $32 \times 32$ image size.
\section{Ablation studies}
In this section, we show ablation studies showing the effect of the components we have added to our architecture. We chose a basic architecture, which does not have any linear block, shifting layer, and only 30 mixer layers, as a baseline and show how different components affect the results. We make all the evaluations on CIFAR-10.

Our experiments in Table~\ref{tab:ablation} confirm that adding linear blocks and shift layers helped improve density estimation by a huge margin. Additionally, our result with 60 layers indicates stronger performance as the number of layers gets larger for our architecture.

\begin{table}[!htbp]
  \caption{Negative log-likelihood (in bits per dimension)
  Smaller values are better.}
  \label{tab:ablation}
  \centering
  \begin{tabular}{ll}
    \toprule
    %\multicolumn{4}{c}{Datasets}                
    %\cmidrule(r){3-7}
    
    Additions & BPD  \\
    \midrule
    Basic &  3.81   \\
    \midrule
    60 layers &  3.52     \\
    Linear blocks & 3.49 \\
    Shift layers & 3.58   \\
    %\cmidrule(r){1-7}
    \bottomrule
  \end{tabular}
\end{table}

\section{Things that did not help}

\paragraph{\textbf{Shift layer with larger shifts:}} In our experimentation with $32\times 32$ datasets, we explored the use of larger shifts and more frequent applications of the shift layer. Whilst these modifications proved to be advantageous compared to not using a shift layer at all, they did not yield great improvements when the shift exceeded a value of 1 for these particular datasets. We believe that the mixing requires many layers and the introduction of larger shifts significantly alters the patch definitions of small-resolution images leading to sub-par results.

\paragraph{\textbf{Stripes, bands, or dilated patches:}} Instead of using simple non-overlapping patches, we experimented with stripes, bands, and dilated patches. However, this did not result in any noticeable improvement.

\paragraph{\textbf{Changing the patch size mid-layers:}} We conducted experiments involving the use of different patch sizes within different layers of our model. Specifically, we experimented with smaller-resolution patches in earlier layers, followed by higher-resolution patches (and vice versa). The intention behind this approach was to enable the model to initially learn broader features and progressively refine specific features, similar to the behaviour of convolutional layers. However, this change in patch size did not lead to any noticeable improvements. Additionally, using dilated patches did not yield any performance enhancements.

\paragraph{\textbf{Squeezing operation:}} Following Glow, we tried the squeeze layers to change the spatial resolution into the channel dimension. Unfortunately, it led the models to overfit easily as the training loss went down but the validation loss went up.

\paragraph{\textbf{Iterative growing of flows:}}
We experimented with deeper architectures by first training only a more shallow architecture.
After several epochs, we added more layers with identity initialisation and repeated the process until the desired architecture depth was achieved.
We also worked with different learning rate sizes for different layers.
However, we found no improvement compared to a simple end-to-end training scheme.

\paragraph{\textbf{Training instabilities:}}
We tried to enhance training stabilities by replacing the affine scale functions -- traditionally an exponential function -- with functions bounded away from 0.
We also tried replacing the Gaussian base distribution with fat-tailed distributions such as a Cauchy distribution or a Gaussian mixture spike and slab model.

\paragraph{\textbf{Positional embedding:}}
In early iterations, we tried incorporating either learnable or sinusoidal positional embeddings as conditioning variables into the flow networks but found them to only add unnecessary complexity.

% \begin{figure}
    
%     \centering
%     \includegraphics[width=1.0\linewidth]{mixer_v2_drawio.png}
%     \caption{MixerFlow architecture}
%     \label{fig:architecture}
% \end{figure}

\begin{figure}

    \centering
    \includegraphics[width=1.0\linewidth]{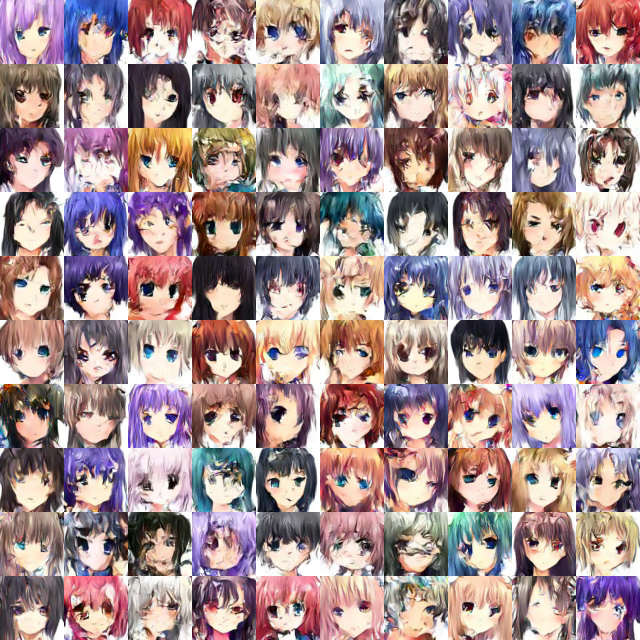} 
    \caption{AnimeFace samples from Glow, Glow has significantly more artefacts than MixerFlow. }
    \label{fig: anime glow full}
\end{figure}

\begin{figure}

    \centering
    \includegraphics[width=1.0\linewidth]{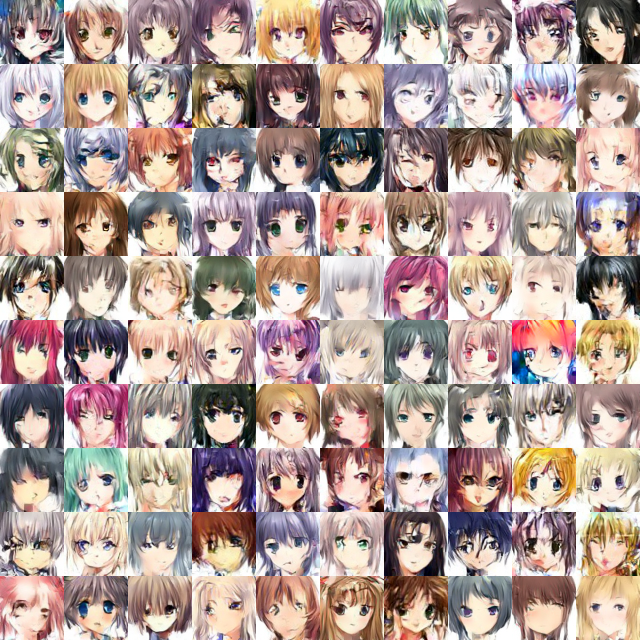} 
    \caption{AnimeFaces samples from our model}
    \label{fig: anime full}
\end{figure}

\begin{figure}

    \centering
    \includegraphics[width=1.0\linewidth]{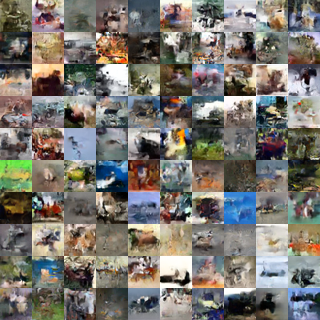} 
    \caption{CIFAR-10 samples from our model}
    \label{fig: cifar10 full}
\end{figure}

\begin{figure}

    \centering
    \includegraphics[width=1.0\linewidth]{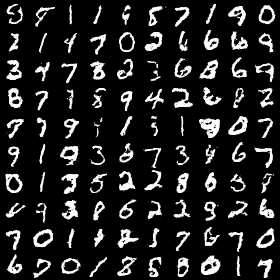} 
    \caption{MNIST samples from our model}
    \label{fig: mnist full}
\end{figure}

\begin{figure}

    \centering
    \includegraphics[width=1.0\linewidth]{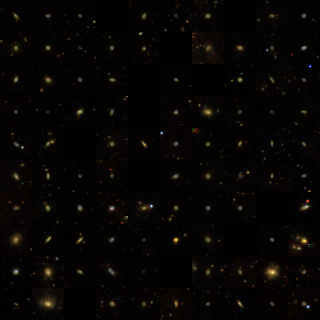} 
    \caption{Galaxy32 samples from our model}
    \label{fig: galaxy full}
\end{figure}

\begin{figure}

    \centering
    \includegraphics[width=1.0\linewidth]{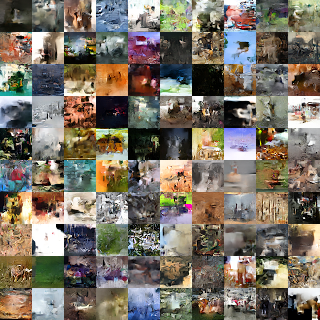} 
    \caption{Imagenet32 samples from our model}
    \label{fig: imagenet32 full}
\end{figure}

\end{document}